\pdfoutput=1

\documentclass[11pt]{article}

\usepackage[final]{acl}

\usepackage{times}
\usepackage{latexsym}

\usepackage[T1]{fontenc}

\usepackage[utf8]{inputenc}

\usepackage{microtype}

\usepackage{inconsolata}

\usepackage{graphicx}

\usepackage{bbm}
\usepackage{amsmath}
\usepackage{booktabs}
\usepackage{algorithm}
\usepackage{algpseudocode}
\usepackage{accsupp}  
\usepackage{multirow}
\usepackage{amssymb}
\usepackage{adjustbox}
\usepackage{makecell}

\title{Token Erasure as a Footprint of Implicit Vocabulary Items in LLMs}

\author{Sheridan Feucht \hspace{0.5em} David Atkinson \hspace{0.5em}  Byron C. Wallace \hspace{0.5em}  David Bau \\
Northeastern University \\
\texttt{\{feucht.s, atkinson.da, b.wallace, d.bau\}@northeastern.edu}}

\begin{document}
\maketitle

\begin{abstract}
LLMs process text as sequences of tokens that roughly correspond to words, where less common words are represented by multiple tokens. However, individual tokens are often semantically unrelated to the meanings of the words/concepts they comprise. For example, Llama-2-7b's tokenizer splits the word ``northeastern'' into the tokens [\texttt{\_n}, \texttt{ort}, \texttt{he}, \texttt{astern}], none of which correspond to semantically meaningful units like ``north'' or ``east.'' 
Similarly, the overall meanings of named entities like ``Neil Young'' and multi-word expressions like ``break a leg'' cannot be directly inferred from their constituent tokens.
Mechanistically, how do LLMs convert such arbitrary groups of tokens into useful higher-level representations? In this work, we find that last token representations of named entities and multi-token words exhibit a pronounced ``erasure'' effect, where information about previous and current tokens is rapidly forgotten in early layers. Using this observation, we propose a method to ``read out'' the implicit vocabulary of an autoregressive LLM by examining differences in token representations across layers, and present results of this method for Llama-2-7b and Llama-3-8b. To our knowledge, this is the first attempt to probe the implicit vocabulary of an LLM.\footnote{Code and data available at \href{https://footprints.baulab.info}{\texttt{footprints.baulab.info}}  }
\end{abstract}

\section{Introduction}
Despite their widespread use, 
the specific mechanisms by which LLMs are able to ``understand'' and generate coherent text are not well understood. 
One mystery is the process by which groups of subword tokens are converted into meaningful 
representations, a process described by \citealp{elhage2022solu} and \citealp{gurnee2023finding} as \textit{detokenization}.

\begin{figure}
    \centering
    \includegraphics[width=\columnwidth]{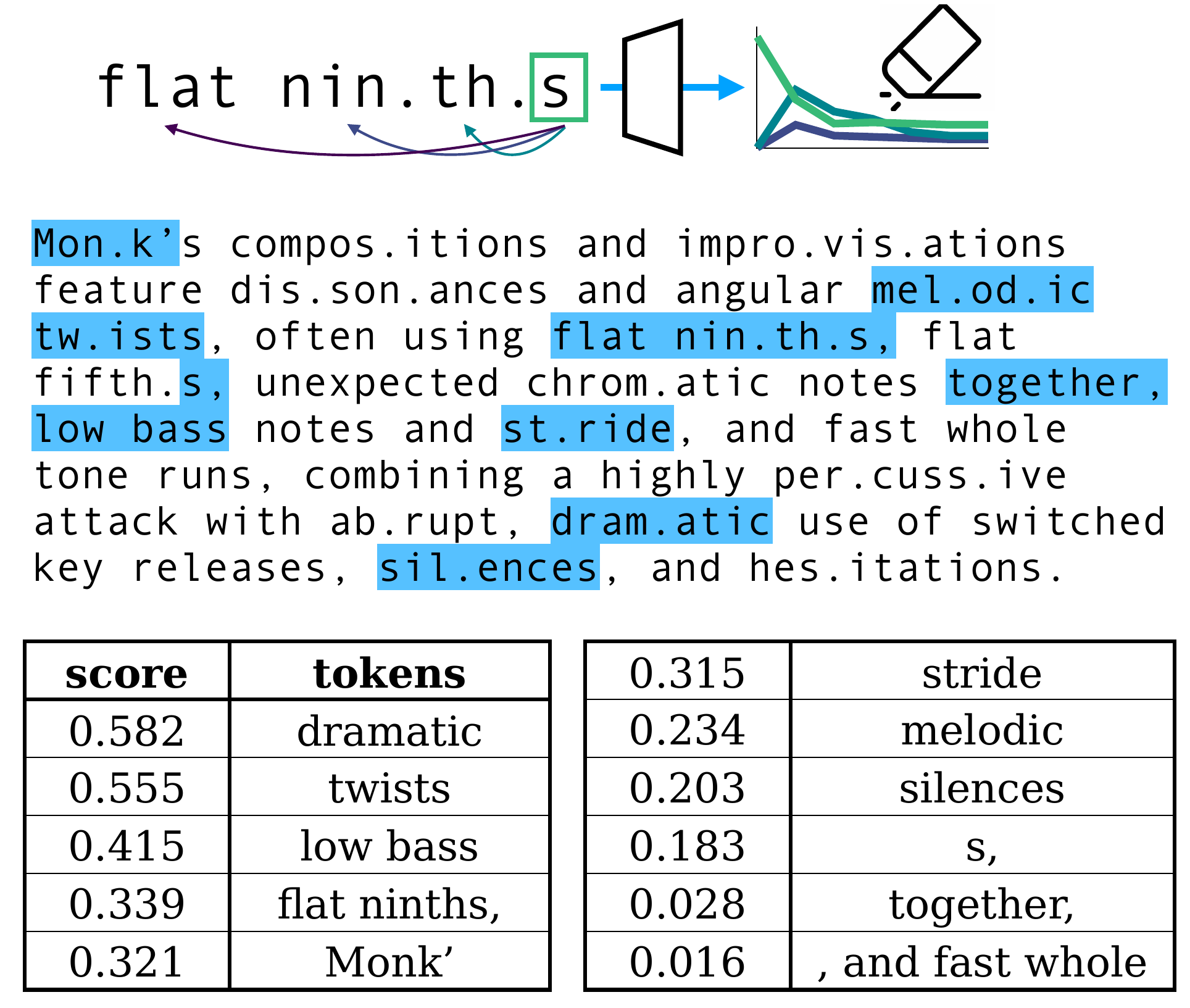}
    \caption{We observe ``erasure'' of token-level information in later layers of LLMs for multi-token words and entities (top). We hypothesize that this is a result of a process that converts token embeddings into useful lexical representations, and introduce a new method for enumerating these lexical items (bottom).
    }
    \label{fig:figure1}
    \vspace{-0.55cm}
\end{figure}

Current language models process text as a series of tokens drawn from a set token vocabulary: One token can correspond to a single word (\texttt{\_fish}), or to a piece of a larger word (\texttt{mon} in ``salmon''). The vocabulary of tokens available to a model is typically determined before training with byte-pair encoding \citep{sennrich-etal-2016-neural}, which is based on a specific dataset and can lead to unintuitive results. For example, Llama-2-7b's \citep{touvron2023llama} tokenizer breaks the word 
``northeastern'' into the tokens [\texttt{\_n}, \texttt{ort}, \texttt{he}, \texttt{astern}], none of which correspond to semantically meaningful units like ``north'' or ``east.'' 
Capitalization also creates unexpected issues: for example, the word ``Hawaii'' is split into two tokens if the first letter is capitalized [\texttt{\_Hawai}, \texttt{i}], but four if the first letter is lowercase [\texttt{\_ha}, \texttt{w}, \texttt{ai}, \texttt{i}]. In spite of these challenges, large models are apparently able to ``understand'' such idiosyncratic tokenizations of multi-token words with few observable effects on downstream performance \citep{gutiérrez2023biomedical}, unless these weaknesses are directly targeted \citep{wang2024tokenization, batsuren2024evaluating}. How is this possible?

We hypothesize that during pretraining, LLMs develop an \textit{implicit vocabulary} that maps from groups of arbitrary tokens to semantically meaningful units. These lexical units may be multi-token words (``northeastern''), named entities (``Neil Young''), or idiomatic multi-word expressions (``break a leg'') and can be understood as ``item[s] that function as single unit[s] of meaning'' in a model's vocabulary \citep{simpson2011routledge}. Lexical items are also non-compositional: Just as the meaning of ``break a leg'' cannot be predicted from the individual meanings of ``break'' and ``leg,'' the meaning of ``patrolling'' cannot be predicted from its constituent tokens \texttt{pat} and \texttt{rolling}.
This arbitrariness necessitates 
some kind of storage system, implicit or otherwise \citep{Murphy_2010}. 

How exactly do LLMs deal with these cases mechanistically? In this paper, we begin to answer this question by investigating token-level information stored in LLM representations.

\begin{itemize}
    \item 
    We find that last token positions of multi-token words and named entities ``erase'' token-level information in early layers for both Llama-2-7b \citep{touvron2023llama} and Llama-3-8b \citep{llama3}.
    \item 
    We develop a heuristic for scoring the ``lexicality'' of a given sequence of tokens, and use it to ``read out'' a list of an LLM's lexical items given a large dataset of natural text. 
\end{itemize}

We interpret this erasure effect as a ``footprint'' of a mechanism in early layers that orchestrates the formation of meaningful lexical items. 

\section{Background}\label{background}

\begin{figure*}
    \centering
    \includegraphics[width=\linewidth]{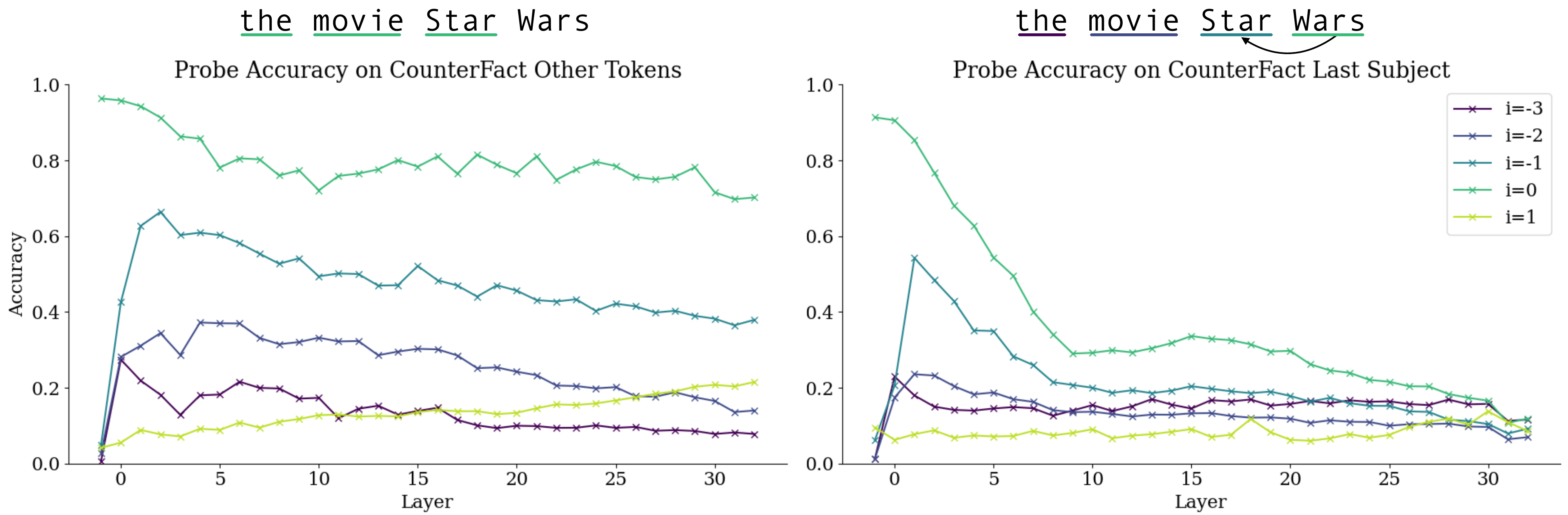}
    \caption{Top-1 test accuracy on \textsc{CounterFact} subject last tokens versus other tokens in the dataset for probes trained on Llama-2-7b hidden states ($n=5063$). $i$ represents the position being predicted (e.g., $i=-1$ is previous token prediction; $i=1$ is next-token prediction). We observe an ``erasure'' effect in last subject tokens that is not present for other types of tokens: these last subject tokens consistently ``forget'' about preceding tokens and themselves. Appendix~\ref{appendix:probes} shows Llama-3-8b results and in-distribution performance on Pile tokens.
    }
    \label{fig:probe-cf}
\end{figure*}

Previous work has shown that knowledge about a multi-token entity is often stored in the last token of that entity. For example, \citet{Meng2022LocatingAE} found that factual information about a subject like ``The Space Needle'' would be concentrated in the representation for \texttt{le}. \citet{Geva2023DissectingRO} find evidence for a \textit{subject enrichment stage} during factual recall, where information about an entity is collected at its last token in early layers, which is also seen in other work on factual recall using the same dataset \citep{katz2024backward}, and corroborated by research on athlete $\rightarrow$ sport lookups \citep{nanda2023factfinding}. This phenomenon may be due to the autoregressive nature of decoder transformer models: models cannot enrich ``Space'' with information about Seattle until after ``Needle'' is seen, as ``Space'' could refer to a number of unrelated concepts (``Space Jam,'' ``Space Station'').\footnote{This is not a hard-and-fast rule; it depends on entity frequency and context cues. For example, if a model sees \texttt{\_The}, \texttt{\_E}, and \texttt{iff}, it may already know that these tokens refer to ``The Eiffel Tower'' without needing to see \texttt{el} and \texttt{Tower}.}  

Other work in interpretability has also started to uncover evidence of models encoding lexical items. \citet{elhage2022solu} observe neurons in early layers that fire on the last tokens of multi-token words, names of famous people, generic nouns, compound words, and {\tt LaTeX} commands. 
They also find late-layer neurons that seem to be relevant to \textit{retokenization}, i.e., conversion from internal representations back into tokens. For example, a retokenization neuron might fire on \texttt{\_st} and promote \texttt{rag} in order to facilitate the output of the word ``straggler.'' \citet{gurnee2023finding} also find examples of polysemantic neurons in Pythia models \citep{mallen2023eliciting} that activate for a number of multi-token constructions like ``apple developer,'' ``Bloom.ington,'' and ``research.gate.''

\section{Linear Probing of Hidden States}\label{probing}

\subsection{Method}\label{method}
If last token positions are so important (Section~\ref{background}), then what do these representations encode? Perhaps the last hidden state directly stores information about other subject tokens (e.g., \texttt{\_Wars} might contain some encoding for \texttt{\_Star} in its hidden state). To test this hypothesis, we investigate hidden states for both Llama-2-7b and Llama-3-8b, as they have significantly different token vocabulary sizes $|\mathcal{V}|$ (32k and 128k tokens, respectively). 

Let $d$ denote the hidden dimension of the model. We train linear probes $p_i^{(\ell)} : \mathbb{R}^d \rightarrow \mathbb{R}^{|\mathcal{V}|}$ to take a hidden state $h_t^{(\ell)}\in \mathbb{R}^d$ at layer $\ell$ and token position $t$ and predict the value of a nearby token $t + i$. For example, a probe trained to predict the previous token for hidden states at layer 5 would be denoted by $p_{-1}^{(5)}$.

We train probes for all layer indexes $0 \leq \ell < 32$ and offsets $i \in \{-3, -2, -1, 0, 1\}$. We also train probes in the same manner on the embedding layer ($\ell=-1$) and on the final outputs of the network before the language modelling head ($\ell=32$). We trained probes on a random sample of 428k tokens from the Pile \citep{pile} using AdamW for 16 epochs with a batch size of 4 and a learning rate of 0.1. Hyperparameters were selected based on validation performance on a separate Pile sample (279k tokens) after a random sweep. Each probe takes 6-8 hours to train on an RTX-A6000.

\subsection{\textsc{CounterFact} Subjects}
After training probes in Section~\ref{method}, we test them on the \textsc{CounterFact} dataset \citep{Meng2022LocatingAE}, which consists of prompts about subjects that require factual knowledge to complete correctly (e.g. ``Mount Passel is in Antarctica''). We filter the dataset to include only prompts that the model answers correctly, yielding 5,063 examples for Llama-2-7b and 5,495 examples for Llama-3-8b. To augment this dataset, we also sampled and filtered down [album/movie/series $\rightarrow$ creator] pairs from Wikidata \cite{wikidata} and embedded them in prompts in the same manner, yielding a total of 12,135 correctly-answered prompts for Llama-2-7b and 13,995 for Llama-3-8b.

Figure~\ref{fig:probe-cf} shows probe test results on \textsc{CounterFact} last subject tokens (right) versus every other type of token in the dataset (left). We see a striking ``erasure'' effect for last tokens of \textsc{CounterFact} subjects, where these hidden states consistently ``forget about'' preceding and current tokens. Subject tokens that are not in the last position (e.g., \texttt{\_Star}) do not exhibit this pattern (Appendix~\ref{appendix:probes}, Figure~\ref{fig:probe-cf-detailed}). This striking drop in token accuracy is reminiscent of the subject enrichment stage described by \citet{Geva2023DissectingRO}, suggesting that the tokens \texttt{\_Star} and \texttt{\_Wars} may be overwritten in the process of representing the concept of \textit{Star Wars}. 

We also observe the same phenomenon when testing on named entities identified by {\tt spaCy} in Wikipedia articles (Appendix~\ref{appendix:probes}, Figure~\ref{fig:probe-mte}), suggesting that this effect is not an artifact of the short templates found in the \textsc{CounterFact} dataset. It also does not seem to be a result of any imbalances in probe training data (Appendix~\ref{appendix:balance}).

\vspace{-0.25em}
\subsection{Multi-Token Words}\label{mtw}
\begin{figure*}[t]
    \centering
    \includegraphics[width=\linewidth]{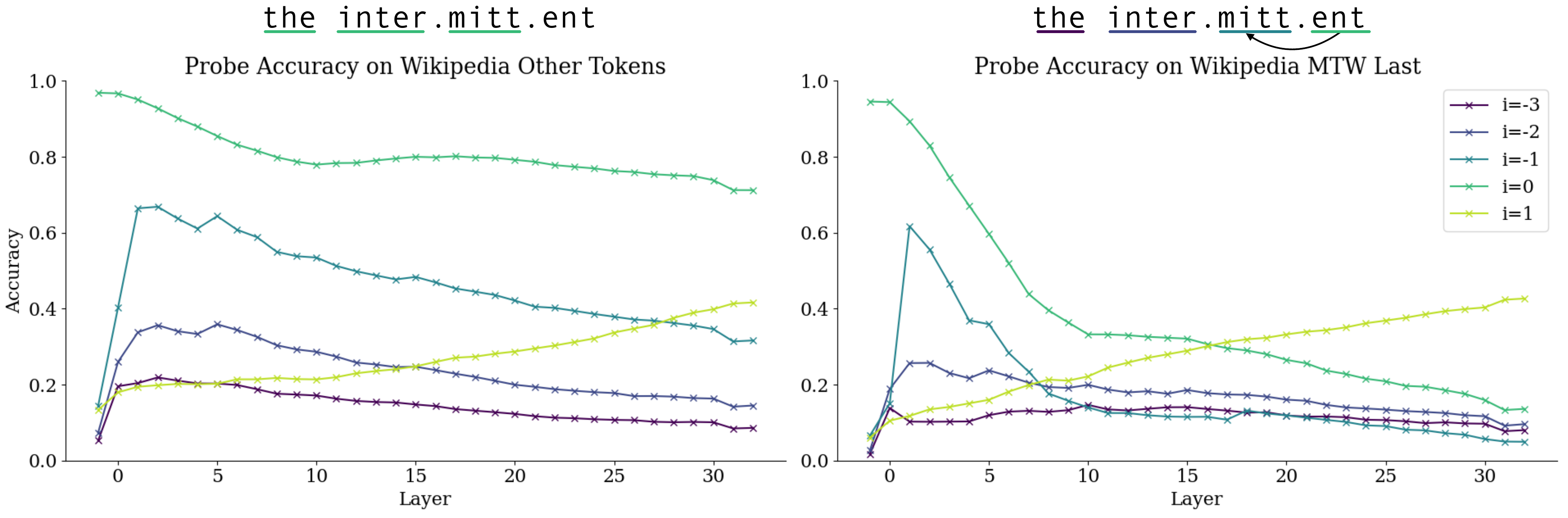}
    \caption{Top-1 test accuracy of probes on last tokens of Wikipedia multi-token words for Llama-2-7b ($n=80606$). Accuracy on all other tokens shown on the left. We see an erasing effect for multi-token words, similar to the effect seen for \textsc{CounterFact} subjects in Figure~\ref{fig:probe-cf}.}
    \label{fig:probe-mtw}
\end{figure*}

Intuitively, the process of converting a multi-token sequence like 
[\texttt{\_n}, \texttt{ort}, \texttt{he}, \texttt{astern}] 
into a meaningful representation of the word ``northeastern'' 
resembles the process of converting [\texttt{\_E}, \texttt{iff}, \texttt{el}, \texttt{Tower}] into ``Eiffel Tower.'' 
We hypothesize that models treat multi-token words in the same way that they treat multi-token entities, and test our probes from Section~\ref{method} on multi-token words. After sampling 500 articles ($\sim$256k tokens) from the \texttt{20220301.en} split of the Wikipedia dump \citep{wikidump}, we split by white-space to naively identify word boundaries. As predicted, we see the same ``erasing'' pattern for multi-token words that we do for multi-token entities (Figure~\ref{fig:probe-mtw}). This suggests that they may be processed in a similar manner in early layers. Appendix~\ref{appendix:probes} shows similar results for Llama-3-8b. 

\vspace{-0.25em}
\section{Building a Vocabulary}\label{building}
After examination of probe behavior for multi-token words and entities, we hypothesize that this ``erasure'' effect is a result of the implicit formation of lexical representations in early layers. To characterize this phenomenon, we propose an \textit{erasure score} $\psi$ to identify token sequences that follow the pattern observed in Section~\ref{probing}. We then introduce an approach to ``reading out'' a list of implicit vocabulary entries for a given model using this score.

\subsection{An Erasure Score}
Given some arbitrary sequence of tokens from indices $p$ through $q$, we want to design an \textit{erasure score} that captures intuitions from Section~\ref{probing}. This score should be higher for sequences exhibiting token erasure (which we hypothesize to be lexical items like [\texttt{\_Cal}, \texttt{g}, \texttt{ary}]), and lower for other types of token sequences (e.g., [\texttt{\_go} \texttt{\_to}, \texttt{\_Cal}, \texttt{g}]).
We design a metric $\psi_{p,q}$ that uses probe outputs from Section~\ref{probing} to look for erasure effects betwen layer 1 and layer $L$.\footnote{For both Llama-2-7b and Llama-3-8b we set $L=9$.}

Concretely, Equation~\ref{eq:erasing-score} defines the score $\psi_{p,q}$ for a sequence $s_{p,q}$ of length $n=q-p+1$ as: 
\begin{equation} \label{eq:erasing-score}
    \frac{1}{1+2n} \left(\delta(q, 0) + \sum_{t=p}^q\sum_{i=-2}^{-1} \mathbbm{1}_{\text{within}}(t,i)\cdot\delta(t,i)\right)
\end{equation}
\noindent where $\delta(t, i)$ denotes the change in probability of the predicted token $t+i$ from layer 1 to layer $L$, based on probes $p_i^{(\ell)}$ from Section~\ref{method}. We take the softmax of the probe outputs to obtain the probability of a specific token $t+i$ in Equation~\ref{eq:delta}. 
\begin{equation}\label{eq:delta}
    \delta(t, i) = P_{p^{(1)}_i}(t+i | h_t^{(1)}) - P_{p^{(L)}_i}(t+i | h_t^{(L)})
\end{equation}
\noindent Finally, if $t+i$ lies outside the boundaries of $s$, we want the score to decrease. If it is within the boundaries of $s$, we want a large drop between layers $\delta(t,i)$ to increase the value of $\psi_{p,q}$. 

\begin{equation}
    \mathbbm{1}_{\text{within}}(t, i) = \begin{cases}
    -1 \text{ if } t + i < p\\
    1 \text{ else}
    \end{cases}
\end{equation}

In summary, for every token position $p\leq t \leq q$ and prediction offset $i \in \{-2, -1\}$, we measure the drop in the predicted probability of the correct token $t+i$ between layer 1 and layer $L$. The more that the probability of the correct answer \textit{decreases} in early layers, the higher we score that sequence. However, if this ``forgetting'' occurs for tokens outside of the boundaries of $s$, we subtract that value from the overall score, effectively penalizing the sequence. This intuition comes from close inspection of probe behavior---for example, Figure~\ref{fig:probe-cf-detailed} shows that there is no ``forgetting'' effect for $i=-1$ when probing the first token of \textsc{CounterFact} subjects. With this approach, we can also account for cases where $s$ is a subsequence of a larger lexical item: if the token \texttt{g} shows a forgetting effect for \texttt{\_Cal} in [\texttt{\_Cal}, \texttt{g}, \texttt{ary}], then the sequence [\texttt{g}, \texttt{ary}] would be penalized. 
Finally, $\delta(q,0)$ additionally rewards sequences in which the last token ``forgets itself,'' as seen in Figures~\ref{fig:probe-cf} and \ref{fig:probe-mtw}. We then normalize by the total number of $\delta$ values considered, in order to account for differing sequence lengths.




\subsection{Segmenting Documents}\label{segmentation}
We develop an algorithm built around our erasure score $\psi$ that breaks any given document $d \in \mathcal{D}$ into high-scoring, non-overlapping segments covering all of $d$ (Algorithm~\ref{algo}). Figure~\ref{fig:figure1} shows the top-scoring sequences $s_{p,q}$ calculated in this manner from a Wikipedia excerpt about Thelonious Monk, where unigram scores are excluded for clarity. Not all multi-token words are scored highly via our approach, but the highest-scoring sequences are plausible lexical items that are non-compositional in nature (``dram.atic'', ``sil.ences'', ``tw.ists''). We share examples of more documents with complete segmentations in Appendix~\ref{appendix:segmentation}. 

\begin{algorithm}
\caption{Document Segmentation}
\begin{algorithmic}[1]
\Require document $d \in \mathcal{D}$ of length $l$
\For{$n = 1$ \textbf{to} $l$} \Comment{all ngram lengths}
    \For{$p = 0$ \textbf{to} $l - n$}
        \For{$q = p + n - 1$ \textbf{to} $l - 1$}
            \State assign score $\psi_{p,q}$ to sequence $s_{p,q}$ 
        \EndFor
    \EndFor
\EndFor
\State sort $s$ in descending order of $\psi$
\State $segms \gets \emptyset$
\For{$s_{p,q}$ in sorted $s$}
    \If{$\forall s_{x,y} \in segms, (x > q \lor y < p)$}
        \State $segms \gets segms \cup \{s_{p,q}\}$
    \EndIf
\EndFor
\State \textbf{return} $segms$ \Comment{non-overlapping segments}
\end{algorithmic}\label{algo}
\end{algorithm}

\subsection{Model Vocabularies}


\begin{table}
\begin{tabular}{lrll}
\toprule
Token Sequence & $n$ & ct & $\psi$ \\
\midrule
lower case & 3 & 2 & 0.736012 \\
storm & 2 & 4 & 0.716379 \\
excursion & 4 & 2 & 0.713134 \\
====... \textit{(72 `equals' signs)} & 8 & 2 & 0.712982 \\
Mom & 3 & 2 & 0.706778 \\
acre & 3 & 2 & 0.629213 \\
Subject & 3 & 2 & 0.607172 \\
ninth & 3 & 2 & 0.606669 \\
processing elements & 3 & 2 & 0.599549 \\
CVC & 3 & 2 & 0.596735 \\
\bottomrule
\end{tabular}
    \caption{Top ten highest-scoring sequences for Llama-2-7b using a Pile subsample (1658 sequences recovered total). $n$ is the number of tokens in the sequence, and `ct' represents occurrences of this segment. $\psi$ is averaged over all occurrences.}
    \label{tab:pile_llama2}
\end{table}

Finally, we propose a method to ``read out'' the implicit vocabulary of a model $\mathcal{M}$ given a dataset $\mathcal{D}$. For each document $d\in\mathcal{D}$, we segment $d$ using Algorithm~\ref{algo}. We then average scores $\psi$ for every multi-token sequence that appears more than once in $\mathcal{D}$. As this process is very data-dependent, we show results for both Pile and Wikipedia text. The top 50 sequences for each dataset and model are provided in Appendix~\ref{appendix:vocab}.  

With this approach, we are able to recover $\sim$1800 sequences for Llama-2-7b and $\sim$900 for Llama-3-8b using the same five hundred Wikipedia articles. Although recall is quite low (Table~\ref{tab:quickeval}), we find that 44.9\% of sequences recovered for Llama-2-7b on Wikipedia text are either multi-token words or multi-token entities (29.68\% for Pile text). For Llama-3-8b, only 5\% and 3\% of retrieved sequences are multi-token words or entities. However, looking at examples of Llama-3-8b sequences in Appendix~\ref{appendix:vocab}, we can observe other interesting cases, like multi-token expressions (``gold medalists,'' ``by per capita income,'' ``thank you for your understanding'') and \texttt{LaTeX} commands (as similarly observed by \citet{elhage2022solu}). 
Because Llama-3-8b's \textit{token} vocabulary is four times larger than Llama-2-7b's, its \textit{implicit} vocabulary also seems to consist of larger multi-word expressions and chunks of code rather than multi-token words (Appendix~\ref{appendix:vocab}, Table~\ref{tab:pile_llama3}).
 
\vspace{-0.25em}
\section{Conclusion}
In this work, we present preliminary evidence for the existence of an \textit{implicit vocabulary} that allows models to convert from byte-pair encoded tokens to useful lexical items. We posit that the ``erasure'' effect we observe for Llama-2-7b and Llama-3-8b is a result of model processes that deal with multi-token expressions, and use this insight to propose a new method for ``reading out'' an LLM's implicit vocabulary. This is a first step towards understanding the formation of lexical representations in LLMs, and may serve as a useful tool for elucidation of words that a given model ``knows.''

\begin{table}
\centering
\begin{tabular}{@{}llllll@{}}
\toprule
                      &      & \multicolumn{2}{c}{MTW} & \multicolumn{2}{c}{MTE} \\ \midrule
llama                 & data & prec.      & recall     & prec.      & recall     \\ \midrule
\multirow{2}{*}{2-7b} & wiki & 0.306      & 0.016      & 0.143      & 0.016      \\
                      & pile & 0.296      & 0.017      & 0.080      & 0.018      \\ \midrule
\multirow{2}{*}{3-8b} & wiki & 0.044      & 0.001      & 0.010      & 0.000      \\
                      & pile & 0.023      & 0.001      & 0.012      & 0.001      \\ \bottomrule
\end{tabular}
\caption{Precision and recall for aggregated results of Algorithm~\ref{algo} run on Llama-2-7b and Llama-3-8b, using either Wikipedia or Pile documents ($|\mathcal{D}|=500$). MTW refers to all multi-token words in the dataset when split by whitespace; MTE refers to all \texttt{spaCy} named entities. }\label{tab:quickeval}
\end{table}

\section*{Limitations}
Evaluation of implicit vocabulary-building methods (Section~\ref{building}) is challenging due to the lack of a known ground-truth. Our approach is motivated by the desire to understand the inner workings of the model being studied, but we have no authoritative reference that distinguishes between situations where a given sequence gets a high $\psi$ value because it is truly treated as a lexical unit by the model, or where it may be due to an error in our methodology. To quantify our results, we have compared the extracted vocbulary to sequences that we assume to be likely lexical items: multi-token words and \texttt{spaCy} named entities. However, this likely does not cover all cases for which ``token grouping'' occurs in LLMs.

Another limitation of this work is that we have restricted our analysis to \textit{known} entities. There is also the question of what happens for intermediate cases such as plausible-sounding fictional towns or names of people who are not famous. If $\psi$ correlates with sequence presence in training data, these results could be useful for understanding how familiar an LLM is with a given word or entity.

Finally, our measurements have been run only on the Llama family of models and do not yet extend to non-Llama models of comparable size, or Llama models of larger sizes.

\section*{Ethics Statement}

In this work, we restrict our analysis to English words, due to our biases as native speakers of English. We hope that this work can also provide valuable insights for other languages, especially low-resource languages, where understanding ``what words an LLM knows'' may be especially useful.

\section*{Acknowledgments}

We thank Koyena Pal, David Smith, Bilal Chughtai, Chantal Shaib, Atticus Geiger, and Adrian Chang for helpful discussion and feedback throughout the course of this project. This work was supported in part by Open Philanthropy, and by the National Science Foundation (NSF) grant IIS-1901117.

Experiments were implemented using the \href{https://nnsight.net/}{\texttt{nnsight}} library \citep{nnsight}. Many were run on the Center for AI Safety Compute Cluster. Any opinions, findings, and conclusions or recommendations expressed in this material are those of the author(s) and do not necessarily reflect the views of the sponsors.

\bibliography{custom}

\clearpage

\appendix

\section{Additional Probing Results}\label{appendix:probes}

\subsection{Llama-3-8b Results}
\paragraph{\textsc{CounterFact} Accuracy} We share results  analogous to Figure~\ref{fig:probe-cf} for Llama-3-8b, which shows a similar ``erasure'' pattern (Figure~\ref{fig:probe-cf-3}). Probes are tested only on prompts that Llama-3-8b answers correctly. 

\paragraph{Multi-Token Word Accuracy}\label{mtw-3} Figure~\ref{fig:probe-mtw-3} shows results for Llama-3-8b probes tested on the last token positions of multi-token words from Wikipedia (where ``words'' are determined by whitespace separation). 

\paragraph{Multi-Token Entity Accuracy} Figure~\ref{fig:probe-mte-3} shows results for probes tested on the last token positions of multi-token entities identified by \texttt{spaCy}, using the same dataset that we do for multi-token words. We use \texttt{spaCy}'s named entity recognition pipeline to identify named entities. Because digits 0-9 are added to Llama-2-7b's vocabulary, we filter out all classes relating to numbers (\texttt{PERCENT}, \texttt{DATE}, \texttt{CARDINAL}, \texttt{TIME}, \texttt{ORDINAL}, \texttt{MONEY}, \texttt{QUANTITY}), with the thought that these sequences may be treated differently at the detokenization stage.

\subsection{Llama-2-7b Results}

\paragraph{Multi-Token Entity Accuracy} Figure~\ref{fig:probe-mte} shows results for Llama-2-7b probes tested on multi-token entities from Wikipedia, using the same dataset from Section~\ref{mtw} and also filtering out number-based entity classes as in Section~\ref{mtw-3}. 

\begin{figure}
    \centering
    \includegraphics[width=\columnwidth]{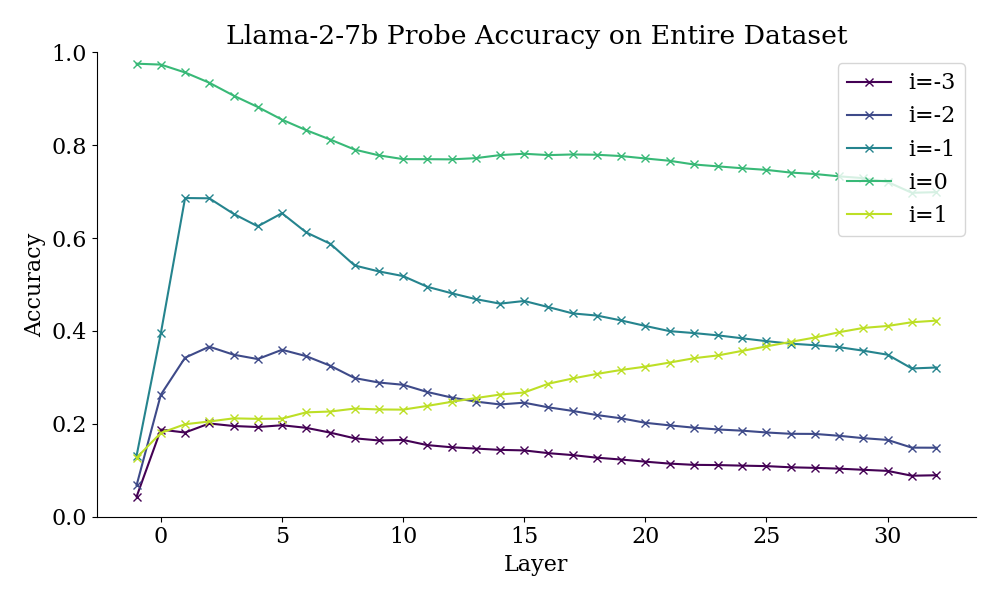}
    \caption{Overall test accuracy on unseen Pile tokens ($n=273$k) for probes trained on Llama-2-7b hidden states. Next token prediction becomes more accurate throughout model layers as current and previous token accuracy decreases.} 
    \label{fig:probe-test}
\end{figure}

\paragraph{Pile Accuracy} While Figure~\ref{fig:probe-cf} shows test accuracy of linear probes on model hidden states, Figure~\ref{fig:probe-test} shows in-distribution test accuracy on Pile tokens. We can observe a smoother trajectory of gradual ``forgetting'' of previous and current token-level information throughout layers. 

\paragraph{Comparison of Token Positions} Figure~\ref{fig:probe-cf-detailed} shows the breakdown of probe performance on different types of subject tokens: first subject tokens, middle subject tokens, and last subject tokens. We see that the observed drop in previous and current token representation observed in last subject tokens still exists, but is not as drastic for first and middle subject tokens. 

\paragraph{Comparison of Subject Lengths} We also show previous token representation broken down by \textsc{CounterFact} subject length for last token representations in Figure~\ref{fig:probe-ngram}. Unigram subjects represent previous token information at a rate even higher than non-subject tokens. For bigrams and trigrams, we see a pattern similar to Figure~\ref{fig:probe-cf}.

\section{Accounting for Possible Training Imbalance}\label{appendix:balance}
One explanation for the observed drop in accuracy for \textsc{CounterFact} entities across layers is that our probes have simply not been exposed to as many entity tokens during training. We do not believe this is the case for Llama-2-7b for two reasons: (1) If this effect was due to probes being less sensitive to tokens found in multi-token entities, we would also see a significant drop for first and middle tokens, which does not occur (Figure~\ref{fig:probe-cf-detailed}). (2) We measure the frequency of all test n-grams in the original Pile data used to train our probes, and find that both subject and non-subject n-grams are found in the probe training dataset at similar rates, with the median number of occurrences in the test set for both types of sequences being zero. After removing the few non-subject sequences that do appear often in the probe training set, we still see the same ``erasure'' effect.

\section{Choice of $L$}

We choose $L=9$ based on probe behavior for Llama-2-7b and Llama-3-8b, particularly in Figures~\ref{fig:probe-cf} and \ref{fig:probe-mtw}. Table~\ref{tab:l_ablation} shows an additional ablation experiment for $L \in \{5, 9, 13, 17, 21\}$.

\begin{table}
\centering
\begin{tabular}{@{}lllll@{}}
\toprule
    & \multicolumn{2}{c}{MTW} & \multicolumn{2}{c}{MTE} \\ \midrule
$L$ & prec.      & recall     & prec.      & recall     \\ \midrule
5   & 0.307      & 0.002      & 0.143      & 0.002      \\
9   & 0.306      & 0.016      & 0.143      & 0.016      \\
13  & 0.328      & 0.003      & 0.169      & 0.003      \\
17  & 0.330      & 0.003      & 0.180      & 0.003      \\
21  & 0.319      & 0.003      & 0.172      & 0.003      \\ \bottomrule
\end{tabular}
\caption{Precision and recall for different values of $L$ for Algorithm~\ref{algo} applied to Llama-2-7b on Wikipedia text. Recall seems to be best for $L=9$, with precision improving by a few points in mid-late layers.}\label{tab:l_ablation} 
\end{table}

\section{Document Segmentation}\label{appendix:segmentation}

\begin{figure}
    \centering
    \includegraphics[width=\columnwidth]{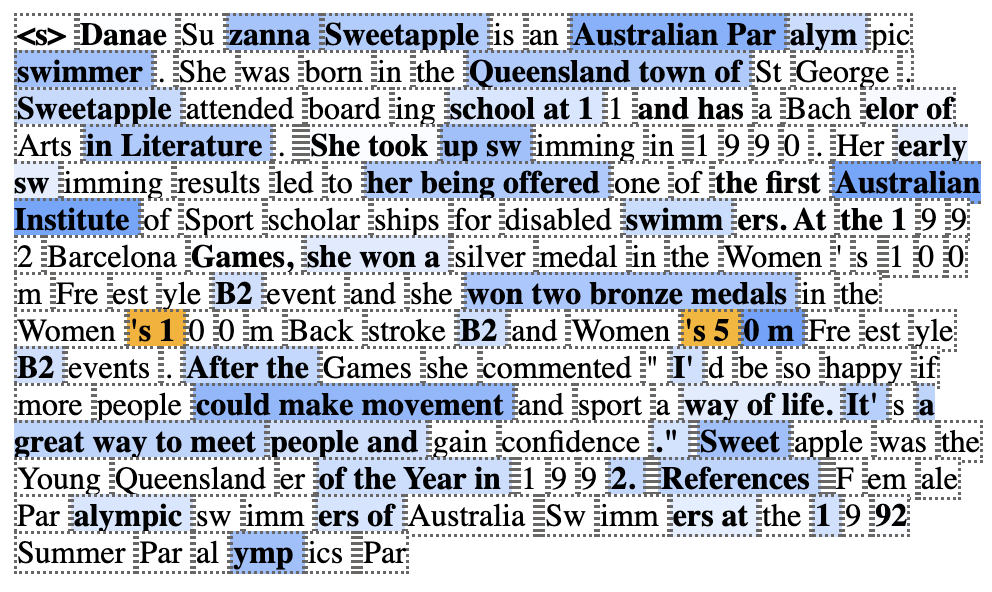}
    \caption{Full segmentation of a document from Wikipedia via Algorithm~\ref{algo} on Llama-2-7b. Borders indicate segmentation, with bolded letters indicating multi-token segments. Darker blue cells have higher scores, yellow cells have negative scores. The highest-scoring sequence in this document is ``Australian Institute'' ($\psi=0.579$).}
    \label{fig:segment-wikillama2}
\end{figure}

\begin{figure}
    \centering
    \includegraphics[width=\columnwidth]{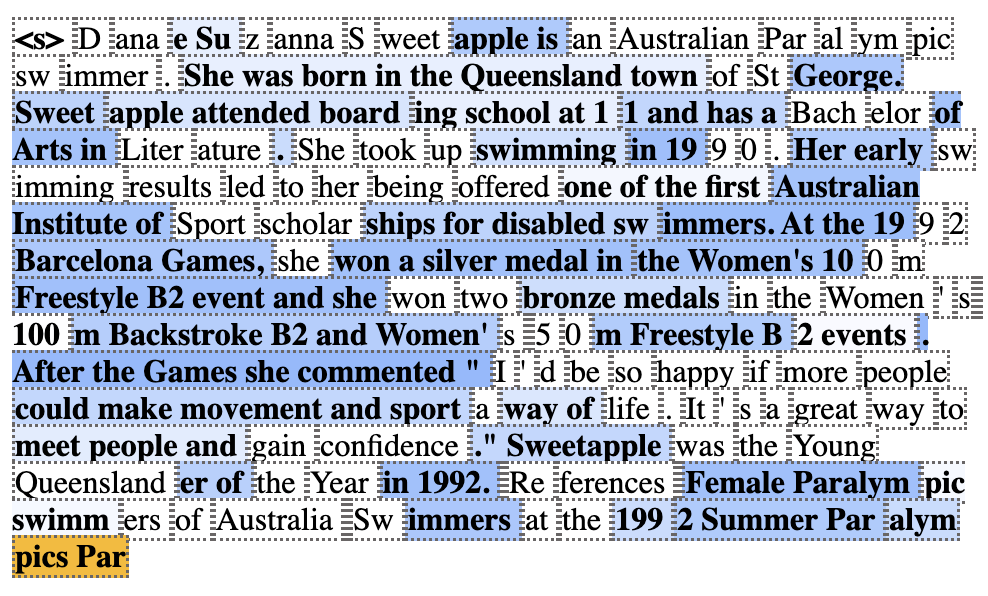}
    \caption{Full segmentation of a document from Wikipedia via Algorithm~\ref{algo} on Llama-3-8b. Borders indicate segmentation, with bolded letters indicating multi-token segments. Darker blue cells have higher scores, yellow cells have negative scores. The highest-scoring sequence in this document is ``. After the Games she commented "'' ($\psi=0.443$).}
    \label{fig:segment-wikillama3}
\end{figure}

\begin{figure}
    \centering
    \includegraphics[width=\columnwidth]{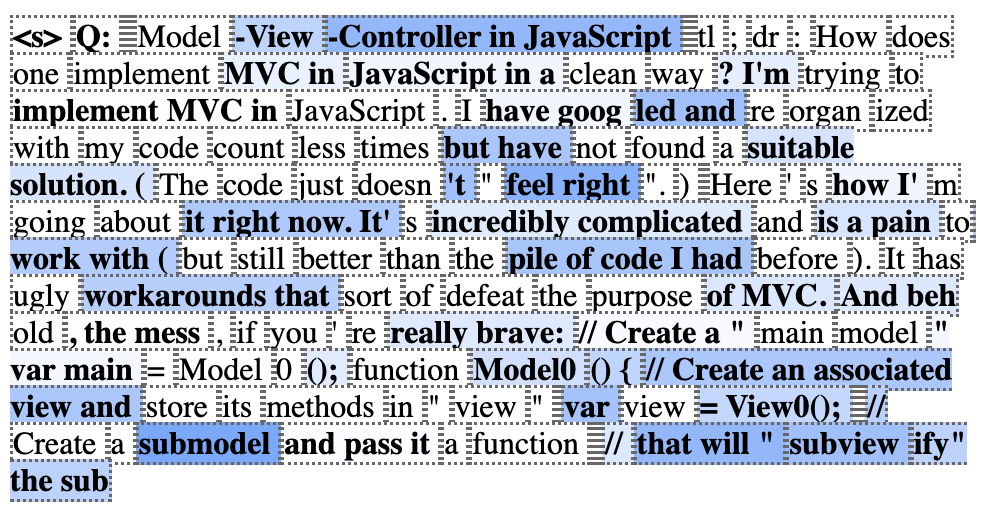}
    \caption{Full segmentation of a document from the Pile via Algorithm~\ref{algo} on Llama-2-7b. Borders indicate segmentation, with bolded letters indicating multi-token segments. Darker blue cells have higher scores, yellow cells have negative scores. The highest-scoring sequence in this document is ``submodel'' ($\psi=0.559$).}
    \label{fig:segment-pilellama2}
\end{figure}

\begin{figure}
    \centering
    \includegraphics[width=\columnwidth]{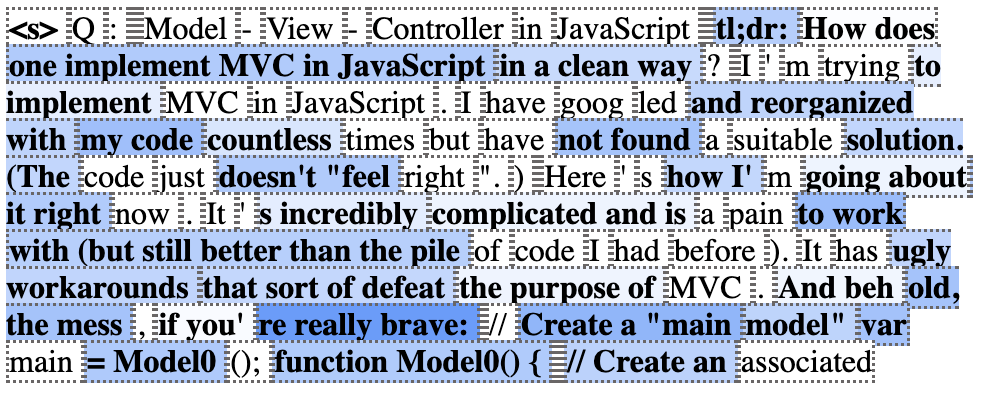}
    \caption{Full segmentation of a document from the Pile via Algorithm~\ref{algo} on Llama-3-8b. Borders indicate segmentation, with bolded letters indicating multi-token segments. Darker blue cells have higher scores, yellow cells have negative scores. The highest-scoring sequence in this document is ``re really brave:'' ($\psi=0.634$).}
    \label{fig:segment-pilellama3}
\end{figure}

\begin{figure*}
    \centering
    \includegraphics[width=\linewidth]{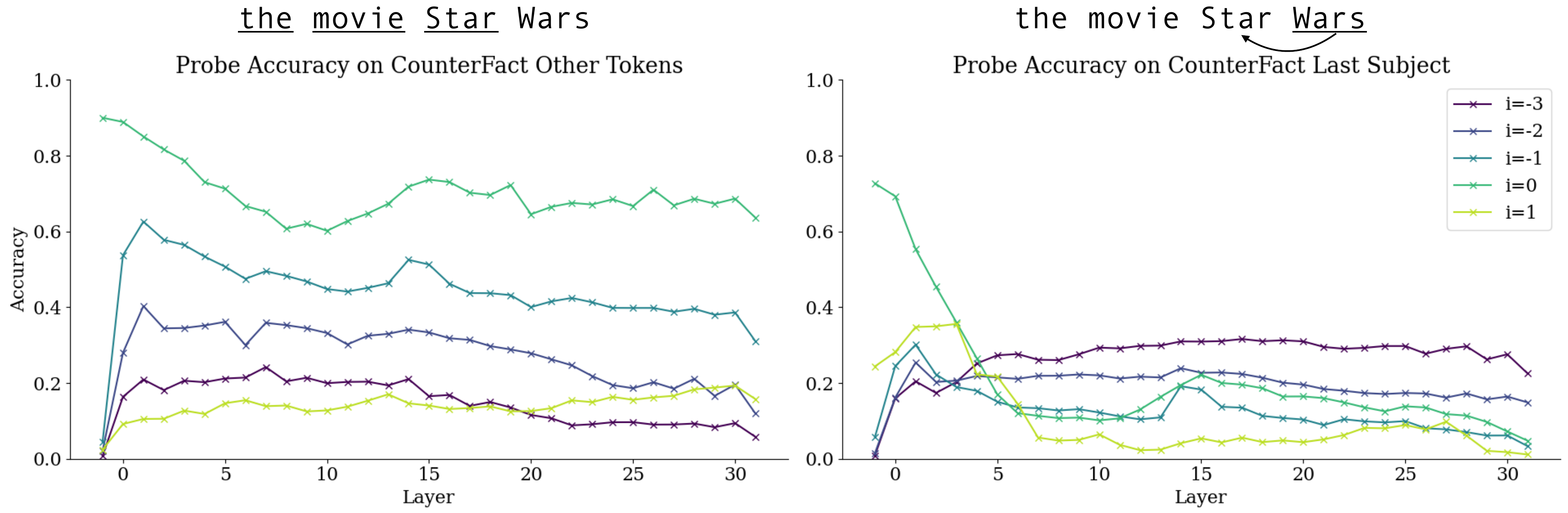}
    \caption{Test accuracy on \textsc{CounterFact} subject last tokens versus other tokens in the dataset for probes trained on \textbf{Llama-3-8b} ($n=5495$). $i$ represents the position being predicted (e.g., $i=-1$ is previous token prediction; $i=1$ is next-token prediction). We observe an ``erasure'' effect similar to Figure~\ref{fig:probe-cf}.
    }
    \label{fig:probe-cf-3}
\end{figure*}

\begin{figure*}[t]
    \centering
    \includegraphics[width=\linewidth]{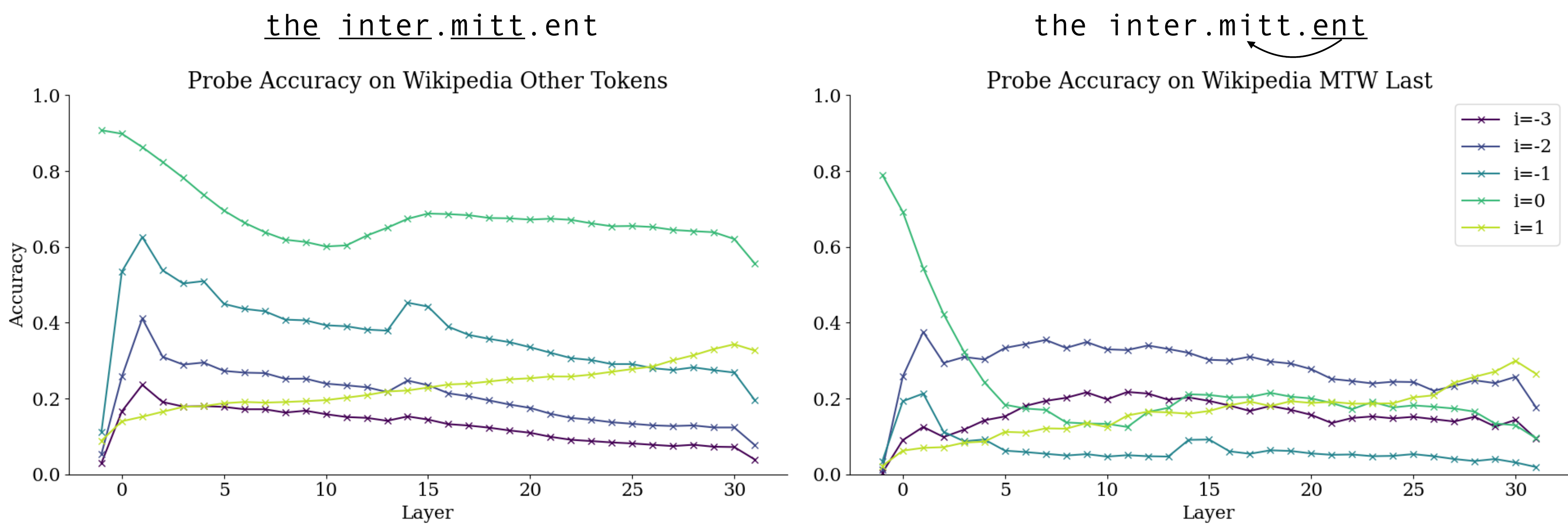}
    \caption{Test accuracy of probes on last tokens of Wikipedia \textbf{multi-token words} for probes trained on \textbf{Llama-3-8b} ($n=91935$; right). Test accuracy on all other tokens shown on the left. Similarly to Figure~\ref{fig:probe-cf}, we see an erasing effect that is not present for other types of tokens.}
    \label{fig:probe-mtw-3}
\end{figure*}

\begin{figure*}[t]
    \centering
    \includegraphics[width=\linewidth]{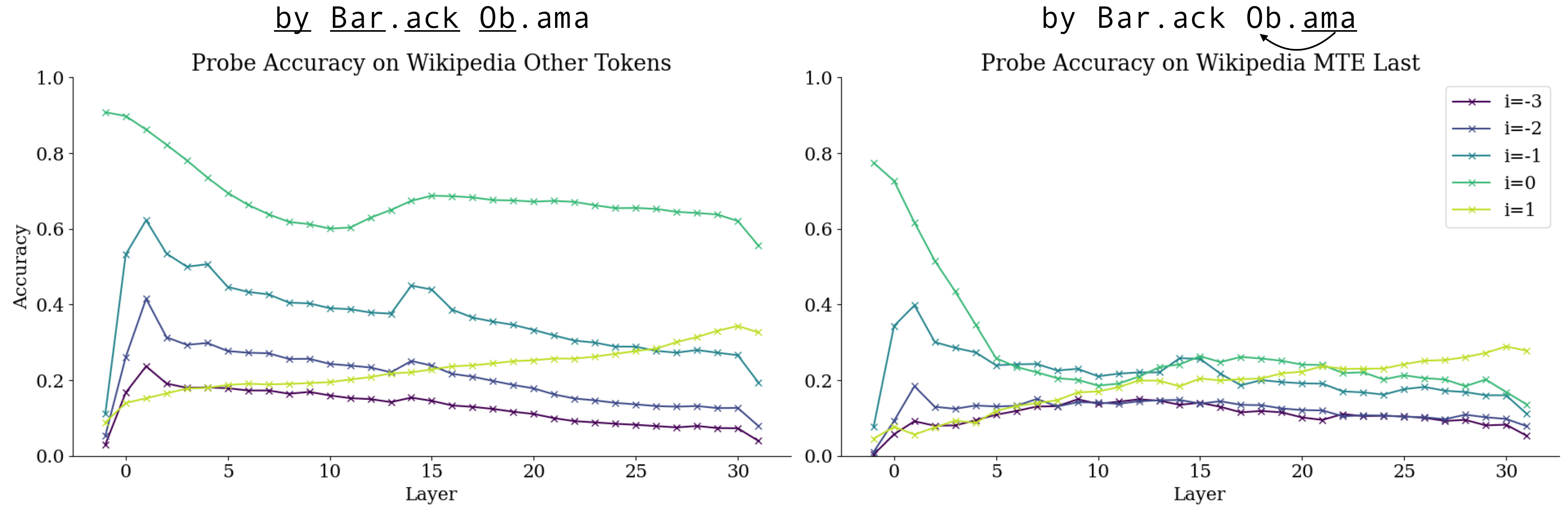}
    \caption{Test accuracy of probes on last tokens of Wikipedia \textbf{multi-token entities} for probes trained on \textbf{Llama-3-8b} ($n=36723$; right). Test accuracy on all other tokens shown on the left. Entities are identified via \texttt{spaCy} named entity recognition, excluding entity types that include digits.}
    \label{fig:probe-mte-3}
\end{figure*}

\begin{figure*}
    \centering
    \includegraphics[width=\linewidth]{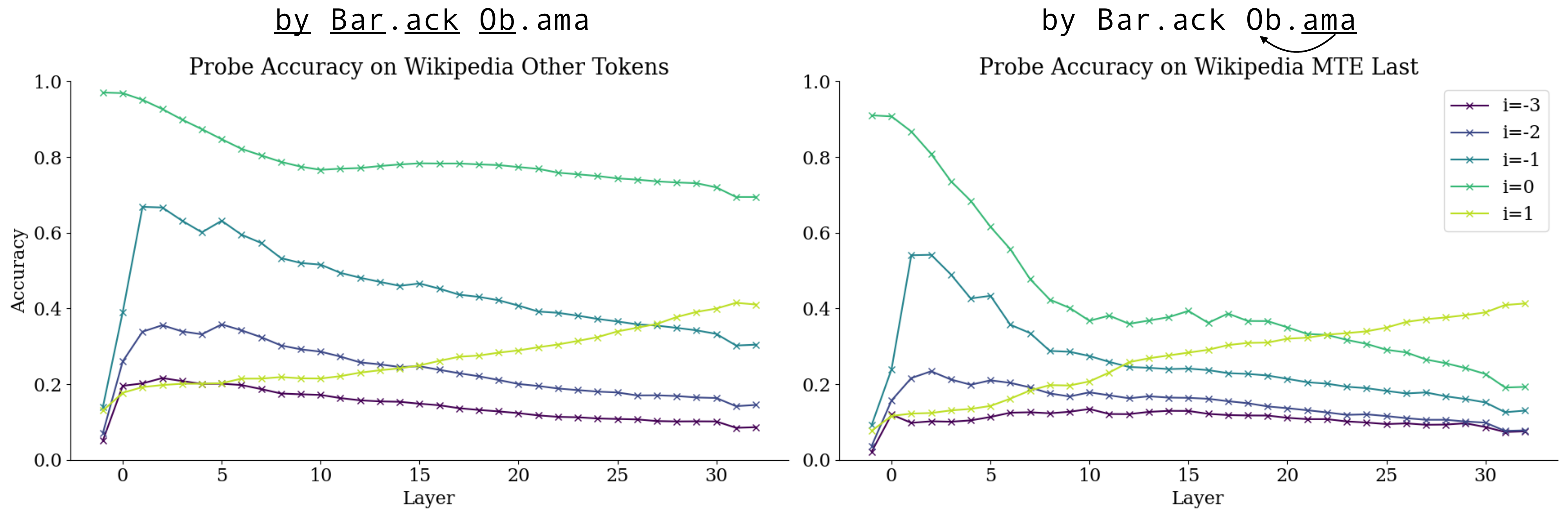}
    \caption{Test accuracy of probes on last tokens of Wikipedia \textbf{multi-token entities} for \textbf{Llama-2-7b} ($n=36723$; right). Test accuracy on all other tokens shown on the left. Entities are identified via \texttt{spaCy} named entity recognition, excluding entity types that include digits.}
    \label{fig:probe-mte}
\end{figure*}

\begin{figure*}
    \centering
    \includegraphics[width=\linewidth]{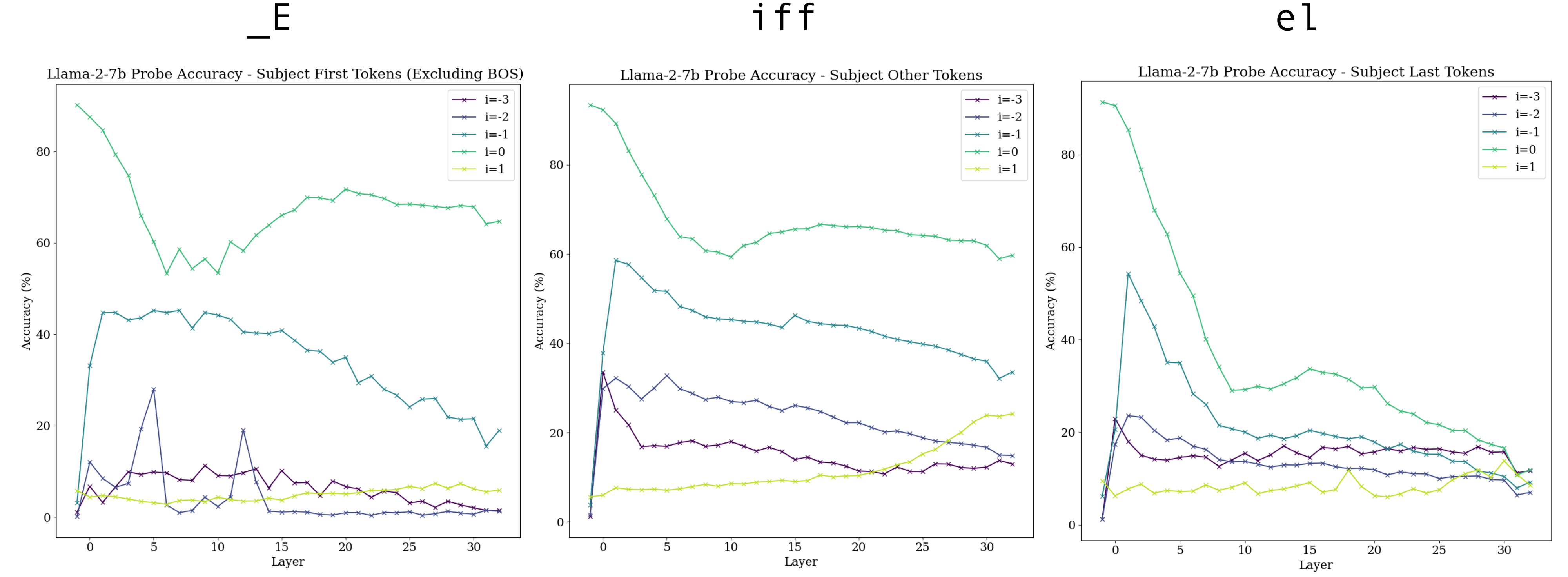}
    \caption{Breakdown for Section~\ref{probing} probes tested on \textsc{CounterFact} first subject tokens, middle subject tokens, and last subject tokens. We observe an ``erasing'' effect only for last subject tokens. Because BOS tokens are recoverable by $i=-1$ probes at high rates, and since 55\% of prompts tested on had subjects at the beginning, we filter examples for which BOS tokens are labels from the leftmost plot.}
    \label{fig:probe-cf-detailed}
\end{figure*}

\begin{figure*}
    \centering
    \includegraphics[width=\linewidth]{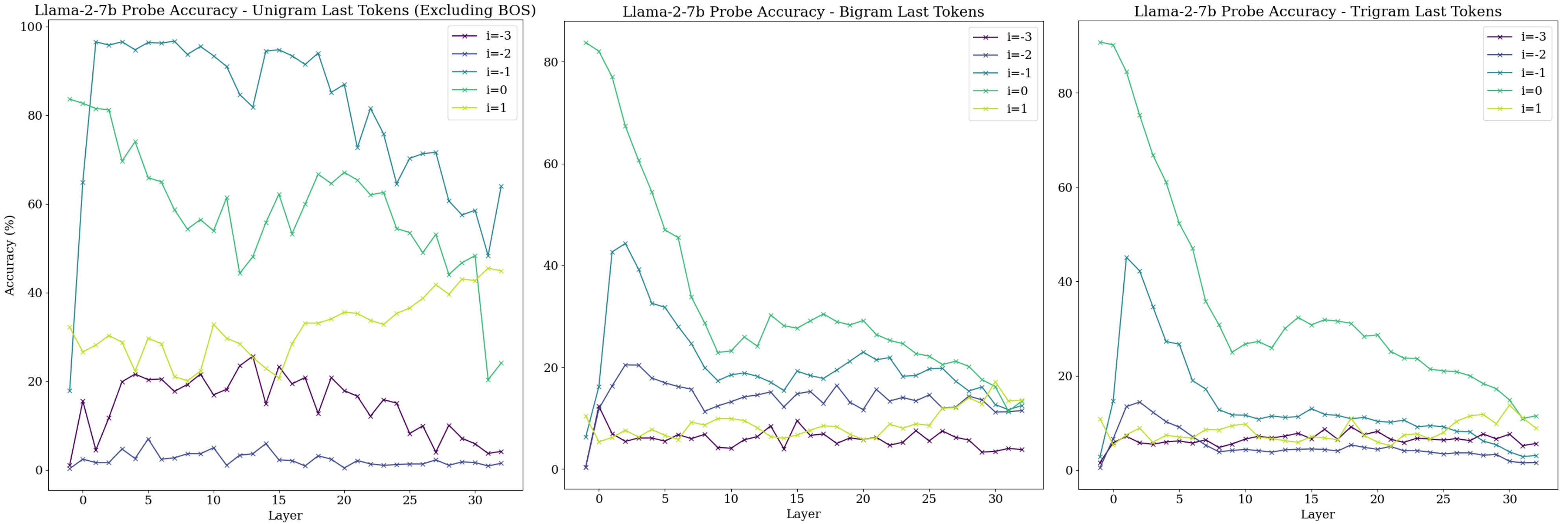}
    \caption{Probe test results for \textsc{CounterFact} subject last tokens broken down for unigrams, bigrams, and trigrams. Unigram subjects store previous token information at rates near 100\%, even excluding BOS tokens.}
    \label{fig:probe-ngram}
\end{figure*}

We provide full document segmentations using Algorithm~\ref{algo} for a short excerpt from a Wikipedia article in Figures \ref{fig:segment-wikillama2} and \ref{fig:segment-wikillama3}. Figures \ref{fig:segment-pilellama2} and \ref{fig:segment-pilellama3} show segmentations for a Pile document.


\section{Model Vocabularies}\label{appendix:vocab}
Tables \ref{tab:wiki_llama2} through \ref{tab:pile_llama3} show the top 50 highest-scoring multi-token sequences for Llama-2-7b and Llama-3-8b across either five hundred Wikipedia articles or five hundred Pile samples. Entries were filtered to show only sequences that appear more than once.



\begin{table}[!t]
\vspace*{-1.5\baselineskip}
\fontsize{11pt}{11pt}\selectfont
    \begin{tabular}{lrll}
    \toprule
    Token Sequence & $n$ & ct & $\psi$ \\
    \midrule
    Gottsche & 3 & 2 & 0.685220 \\
    berth & 3 & 2 & 0.680793 \\
    carries & 3 & 2 & 0.647844 \\
    Eurocop & 3 & 2 & 0.644104 \\
    franchises & 3 & 2 & 0.642707 \\
    0 Women & 3 & 2 & 0.639162 \\
    rape & 3 & 2 & 0.632567 \\
    Rebell & 3 & 3 & 0.614295 \\
    intermittently & 4 & 2 & 0.613479 \\
    enn State & 4 & 3 & 0.607535 \\
    North Dakota & 4 & 10 & 0.600616 \\
    Sride & 3 & 2 & 0.600013 \\
    fiction & 2 & 2 & 0.599339 \\
    Sox & 3 & 3 & 0.599043 \\
    Bazz & 3 & 2 & 0.598242 \\
    erect & 3 & 2 & 0.597915 \\
    borough & 3 & 3 & 0.596054 \\
    encompasses & 5 & 2 & 0.592084 \\
    northernmost & 3 & 2 & 0.591607 \\
    Madras & 3 & 2 & 0.590394 \\
    hull & 3 & 2 & 0.586968 \\
    iron & 2 & 2 & 0.586959 \\
    Galaxy & 3 & 2 & 0.585879 \\
    began operations & 3 & 2 & 0.584680 \\
    Redding & 3 & 2 & 0.584244 \\
    gloss & 3 & 2 & 0.576740 \\
    cello & 3 & 2 & 0.573732 \\
    Gators & 3 & 5 & 0.573675 \\
    senator & 3 & 2 & 0.572947 \\
    restructuring & 4 & 2 & 0.570552 \\
    supervised & 3 & 3 & 0.570421 \\
    Mediterranean & 4 & 2 & 0.567790 \\
    Madera & 3 & 2 & 0.567563 \\
    sequel & 3 & 2 & 0.563626 \\
    scarp & 3 & 3 & 0.561548 \\
    Sout & 3 & 2 & 0.560640 \\
    South Division & 3 & 2 & 0.558720 \\
    rectangular & 3 & 2 & 0.557339 \\
    Danny & 3 & 2 & 0.556836 \\
    Examiner & 4 & 2 & 0.555797 \\
    Kuwait & 4 & 4 & 0.554636 \\
    Bogue & 3 & 6 & 0.552219 \\
    Lancaster & 3 & 3 & 0.552166 \\
    Leuven & 4 & 3 & 0.548806 \\
    the Park & 3 & 2 & 0.548687 \\
    first Baron & 3 & 2 & 0.547447 \\
    fights & 3 & 2 & 0.547171 \\
    Carpio & 3 & 2 & 0.547116 \\
    Czech Republic & 3 & 2 & 0.546651 \\
    Survive & 4 & 2 & 0.546255 \\
    \bottomrule
    \end{tabular}
    \caption{\textbf{Llama-2-7b} Wikipedia results (1808 sequences total). $n$ is the number of tokens in the sequence, and `ct' represents occurrences of this segment. $\psi$ is averaged over all occurrences.}
    \label{tab:wiki_llama2}
\end{table}

\begin{table}[!t]
    \fontsize{11pt}{11pt}\selectfont
    \begin{tabular}{lrll}
    \toprule
    Token Sequence & $n$ & ct & $\psi$ \\
    \midrule
    1992 births & 7 & 2 & 0.573 \\
    19th-century & 7 & 3 & 0.569 \\
     dehydrogen & 5 & 2 & 0.553 \\
     Swahili & 4 & 4 & 0.539052 \\
     Chuck Liddell & 6 & 2 & 0.537169 \\
     its population was  & 5 & 5 & 0.534977 \\
     by per capita income & 6 & 3 & 0.518991 \\
     are brownish & 4 & 2 & 0.515703 \\
    ate women's football & 7 & 4 & 0.509384 \\
     Almeida & 4 & 5 & 0.507277 \\
     of New South Wales & 5 & 3 & 0.503120 \\
    
    2015 deaths & 8 & 2 & 0.503074 \\
    Pittsburgh & 3 & 3 & 0.503070 \\
    
    21st-century & 7 & 4 & 0.499362 \\
     (NSW & 4 & 9 & 0.497107 \\
    age of the United \\ Kingdom & 6 & 3 & 0.487303 \\
    Presidential & 3 & 2 & 0.485317 \\
     Landmark & 3 & 2 & 0.484965 \\
     Alistair & 4 & 2 & 0.484930 \\
     Tauri & 3 & 8 & 0.482449 \\
    2 km & 4 & 2 & 0.479984 \\
    
    20th-century & 7 & 3 & 0.475703 \\
     East Bay & 3 & 2 & 0.475156 \\
     \makecell{game goes in extra \\ time, if the scored} & 10 & 2 & 0.472323 \\
     São Paulo & 3 & 2 & 0.470874 \\
     Atlantic City & 3 & 2 & 0.470726 \\
     Chaluk & 3 & 2 & 0.467165 \\
     Frank Lloyd & 3 & 2 & 0.462585 \\
     may refer to:
     & 6 & 4 & 0.462234 \\
     gold medalists & 4 & 2 & 0.458494 \\
    , 2nd Baron & 6 & 2 & 0.456996 \\
     people)
     & 4 & 4 & 0.454926 \\
     series aired & 4 & 2 & 0.453057 \\
     Srib & 3 & 2 & 0.451708 \\
     with blackish & 4 & 2 & 0.450033 \\
     World Cup players & 4 & 2 & 0.448979 \\
     main role & 3 & 2 & 0.448569 \\
    Bos & 4 & 2 & 0.448425 \\
     Asenath & 4 & 2 & 0.448259 \\
     Royal Navy & 3 & 3 & 0.445617 \\
    2. Bundesliga players & 7 & 2 & 0.445210 \\
    External links & 3 & 69 & 0.444921 \\
     an unincorpor & 6 & 2 & 0.443527 \\
    Gast & 2 & 4 & 0.437695 \\
    Pfor & 3 & 2 & 0.432194 \\
     Elisio de Med & 5 & 2 & 0.431518 \\
    " (2007)
     "Jad & 12 & 2 & 0.429412 \\
     Elkh & 3 & 2 & 0.428984 \\
     Früh & 3 & 2 & 0.427781 \\
     order of the NK & 5 & 2 & 0.424037 \\
    \bottomrule
    \end{tabular}
    \caption{\textbf{Llama-3-8b} Wikipedia results (892 sequences total). $n$ is the number of tokens in the sequence, and `ct' represents occurrences of this segment. $\psi$ is averaged over all occurrences.}
    \label{tab:wiki_llama3}

\end{table}

\begin{table}[t]
\vspace*{-1.9\baselineskip}
\fontsize{11pt}{11pt}\selectfont
\begin{tabular}{lrll}
\toprule
Token Sequence & $n$ & ct & $\psi$ \\
\midrule
lower case & 3 & 2 & 0.736012 \\
storm & 2 & 4 & 0.716379 \\
excursion & 4 & 2 & 0.713134 \\
====... \textit{(72 `equals' signs)} & 8 & 2 & 0.712982 \\
Mom & 3 & 2 & 0.706778 \\
acre & 3 & 2 & 0.629213 \\
Subject & 3 & 2 & 0.607172 \\
ninth & 3 & 2 & 0.606669 \\
processing elements & 3 & 2 & 0.599549 \\
CVC & 3 & 2 & 0.596735 \\
VPN & 3 & 3 & 0.596052 \\
Regul & 3 & 2 & 0.591968 \\
bore & 2 & 2 & 0.590212 \\
\$\textbackslash dot\{G & 5 & 2 & 0.589714 \\
Rates & 3 & 2 & 0.589637 \\
INSURANCE & 5 & 2 & 0.584323 \\
Commercial & 4 & 2 & 0.581543 \\
Barney & 3 & 3 & 0.574872 \\
PTA & 3 & 2 & 0.571932 \\
penetrated & 4 & 2 & 0.570164 \\
MG & 3 & 2 & 0.569830 \\
Leigh & 3 & 2 & 0.567894 \\
jail & 3 & 3 & 0.567225 \\
TNS & 3 & 2 & 0.567003 \\
peptides & 4 & 2 & 0.565775 \\
John Arena & 3 & 2 & 0.565648 \\
Disease & 4 & 2 & 0.564662 \\
welfare & 4 & 4 & 0.564364 \\
wild type & 3 & 2 & 0.560699 \\
uws & 3 & 3 & 0.557799 \\
ongrel & 4 & 3 & 0.554208 \\
liquid cry & 3 & 3 & 0.553408 \\
princess & 3 & 2 & 0.551672 \\
Denmark & 3 & 2 & 0.548702 \\
birthday & 3 & 2 & 0.548504 \\
atedmes & 4 & 2 & 0.548171 \\
"ENOENT & 5 & 2 & 0.547169 \\
third-party & 4 & 2 & 0.546949 \\
aliens & 3 & 2 & 0.546507 \\
Durban & 3 & 4 & 0.545848 \\
Bouncy & 4 & 3 & 0.545826 \\
CHO & 3 & 2 & 0.542762 \\
unjust & 3 & 2 & 0.538813 \\
these motivational & 4 & 3 & 0.537485 \\
DLS & 3 & 4 & 0.535933 \\
\textbackslash n\& & 3 & 2 & 0.534510 \\
uneven & 3 & 2 & 0.533137 \\
watt & 3 & 2 & 0.532243 \\
'She & 3 & 2 & 0.531300 \\
HP & 3 & 3 & 0.529555 \\
\bottomrule
\end{tabular}
    \caption{\textbf{Llama-2-7b} Pile results (1658 sequences total). $n$ is the number of tokens in the sequence, and `ct' represents occurrences of this segment. $\psi$ is averaged over all occurrences.}
    \label{tab:pile_llama2}
\end{table}

\begin{table}[t]
\fontsize{11pt}{11pt}\selectfont
\begin{tabular}{lrll}
\toprule
Token Sequence & $n$ & ct & $\psi$ \\
\midrule
</td>\textbackslash n<td> & 9 & 2 & 0.627583 \\
\{d\}x & 5 & 3 & 0.599395 \\
*\textbackslash n & 4 & 3 & 0.587016 \\
\_\{n=1\}\^\{\textbackslash in & 7 & 4 & 0.585434 \\
</td>\textbackslash n<td & 8 & 2 & 0.573310 \\
-2-2007-061 & 12 & 3 & 0.551581 \\
 reticulum & 4 & 3 & 0.549337 \\
 INSURANCE & 5 & 2 & 0.548263 \\
32;\textbackslash n internal static & 8 & 2 & 0.547893 \\
;\textbackslash n internal static & 6 & 9 & 0.540374 \\
:  At & 4 & 2 & 0.538609 \\
 (2,9,' & 6 & 4 & 0.537495 \\
 Respondent & 4 & 2 & 0.534509 \\
\textbackslash t\textbackslash t\}\textbackslash n\textbackslash n\textbackslash t & 7 & 3 & 0.530669 \\
 (3,0,' & 6 & 4 & 0.529493 \\
\_\{n-1\}\textbackslash ar & 7 & 2 & 0.527303 \\
 \makecell{thank you for \\ your understanding} & 6 & 2 & 0.513979 \\
 hydroxyl & 4 & 2 & 0.510059 \\
>\textbackslash n*/\textbackslash private \$ & 9 & 2 & 0.510054 \\
in mukaan & 5 & 2 & 0.506333 \\
\{w\}\^\{B\}\_\{ & 6 & 2 & 0.505970 \\
/2\textbackslash Z & 5 & 2 & 0.501998 \\
'); \textbackslash nINSERT INTO & 6 & 10 & 0.501055 \\
7-f131 & 7 & 2 & 0.496881 \\
0, 1L>{} & 8 & 2 & 0.495809 \\
/0    S & 5 & 2 & 0.492042 \\
5 Audi & 4 & 2 & 0.491043 \\
 all that apply & 4 & 3 & 0.490469 \\
": true,\textbackslash n & 6 & 2 & 0.486807 \\
4,\textbackslash n& 5 & 2 & 0.485315 \\
 to as DSP & 5 & 2 & 0.484967 \\
**B**]\{\}\textbackslash & 6 & 2 & 0.483484 \\
;\textbackslash ninternal & 5 & 3 & 0.479777 \\
100\% used & 6 & 2 & 0.475673 \\
", "x": & 5 & 3 & 0.474701 \\
2.7 & 4 & 2 & 0.473720 \\
</td>\textbackslash n& 6 & 2 & 0.473578 \\
" code=" & 4 & 4 & 0.473514 \\
e2d-d & 6 & 2 & 0.473418 \\
 is under conversion & 4 & 5 & 0.473355 \\
 \{ int|sys & 5 & 3 & 0.471213 \\
\makecell{();\textbackslash n\}\textbackslash n\textbackslash nprivate \\ boolean isAny} & 12 & 2 & 0.470941 \\
 (2,8,' & 6 & 4 & 0.470214 \\
 trachea & 4 & 2 & 0.469154 \\
 use in an automobile & 6 & 2 & 0.467788 \\
 at org.apache.c & 7 & 5 & 0.467637 \\
 world around us & 4 & 2 & 0.464469 \\
2\textbackslash left(1+x & 8 & 2 & 0.463555 \\
 or Commodore & 5 & 3 & 0.463106 \\
11-117 & 7 & 2 & 0.459824 \\
\bottomrule
\end{tabular}
    \caption{\textbf{Llama-3-8b} Pile results (819 sequences total). $n$ is the number of tokens in the sequence, and `ct' represents occurrences of this segment. $\psi$ is averaged over all occurrences.}
    \label{tab:pile_llama3}
\end{table}

\end{document}